%% file: main.tex
\documentclass[conference]{IEEEtran}

\usepackage{booktabs}
\usepackage{tabularx}
\usepackage{multirow}

\usepackage{graphicx} % Required for inserting images
\usepackage{biblatex} % Imports biblatex package
\usepackage{subcaption}
\usepackage{xcolor}
\title{Comparative Analysis of GAT and BERT for Human-Like Playtesting}

\author{\IEEEauthorblockN{Kleio Fragkedaki, Theodoros Panagiotakopoulos, Matteo Biasielli, Hui Wang}
\IEEEauthorblockA{\textit{AI Center of Excellence} \\
\textit{King}\\
Stockholm, Sweden \\
\{claire.fragkedaki, theodoros.panagiotakopoulos, matteo.biasielli, maddy.hui.wang\}@king.com}
}
\date{February 2025}

\begin{document}

\maketitle

\begin{abstract}
\input{src/0_abstract}
\end{abstract}

\section{Introduction}\label{sec:intro}
\input{src/1_introduction}

\section{Related Works}\label{sec:related-works}
\input{src/1_related_works}

\section{Methodology}\label{sec:methodology}
\input{src/2_methodology}

\section{Results}\label{sec:results}
\input{src/3_results}

\section{Conclusion and Future Work}\label{sec:conclusion}
\input{src/5_conclusion}

\printbibliography 

\end{document}

%% file: src/0_abstract.tex
Accurately modeling and understanding player experience is crucial for designing engaging puzzle games. 
To achieve this, a common approach involves collecting diverse user data to train predictive playtesting models that mimic player behavior. However, existing data-driven methods often lack the ability to capture the full range of player strategies and require extensive feature engineering and network architecture modeling. This limitation becomes particularly evident when new game mechanics or features are introduced, which necessitate continual adjustments to the models.
To address these challenges, we propose a more generalized representation that reduces — or even eliminates — the need for ongoing feature-engineering maintenance. Specifically, we investigate two general-purpose network architectures: (a) a transformer-based model (BERT) and (b) a graph attention model (GAT), both of which are designed to effectively capture the relational structure of Candy Crush Saga (CCS) game boards. Our experiments compare these approaches to Convolutional Neural Networks (CNN) baselines, revealing better performance on challenging board configurations and underscoring the benefits of our generalizable representation.

%% file: src/1_introduction.tex
% 1st paragraph - Problem [what is the problem - to keep users engaged we need to ensure quality of the game etc.] [why playtesting and why human playtesting does not work well]

% Automated playtesting has become an invaluable tool in game development, allowing developers to assess game difficulty, refine level design, and predict player behavior without extensive human testing. By leveraging machine learning models, playtesting can be performed at scale, providing valuable insights into game balance and player experience. 
In game development, balancing difficulty is crucial for player retention and engagement. Challenge plays a fundamental role in player experience, and research indicates that players quickly abandon games that fail to provide an appropriate level of difficulty~\cite{denisova2017challenge}. To optimize gameplay, developers traditionally rely on playtesting to evaluate new levels and fine-tune game parameters before release~\cite{ramadan2013game}. However, human playtesting is costly, time-consuming, and often fails to capture the diversity of player behaviors~\cite{playtesting_beyond_personas}. As a result, machine learning-powered automated playtesting has become a scalable and efficient solution, offering a rapid, data-driven analysis of gameplay across various player profiles and play styles~\cite{gudmundsson2018human, holmgard2019evolvedmcts}.
% In this context, \textit{human-like behavior} refers to a model's ability to replicate player decision-making patterns. We quantitatively assess this alignment using move prediction accuracy, which compares model-selected moves against players’ actual choices in the historical gameplay data. In addition, we evaluate level difficulty alignment by comparing model performance to player success rates, ensuring that the model reflects not only player decisions but also their overall gameplay outcomes.
In this framework, \textit{human-like behavior} refers to the ability of a model to replicate player decisions. This is evaluated using two criteria: (a) move prediction accuracy — how often the model’s actions match human choices in historical data~\cite{shao2016cnn, mcilroy2020aligning} — and (b) level difficulty alignment — how closely model performance correlates with aggregated player success rates~\cite{gudmundsson2018human}.

% 2nd paragraph - Solution of CNN paper in automated playtesting. CNN and some references to other alternatives.
In grid-based puzzle games like Candy Crush Saga (CCS), players generally swap items on a two-dimensional board to create matches based on specific patterns, aiming to accomplish predefined objectives. Due to the spatial layout of these games, Convolutional Neural Networks (CNNs)~\cite{NIPS2012_c399862d} are a popular choice for modeling player interactions as they excel at extracting features from grid-like data~\cite{clark15cnn, shao2016cnn}, 
% Although 
and have been shown to be computationally efficient in game playtesting compared to methods like Monte-Carlo Tree Search (MCTS)~\cite{gudmundsson2018human}.
% , as game mechanics evolve, CNN architectures encounter limitations when dealing with variations in board size and the introduction of new elements and mechanics. 
However, CNN architectures often struggle to generalize when the board size changes or new game mechanics are introduced,
partly due to the convolutional architecture~\cite{azulay2019deep} and partly because their grid-based representation does not reflect the underlying relational topology of the level~\cite{keller2023imagesconnectionsdqngnns}.
% partly due to its architectural constraints such as fixed input dimensions and position-specific filters

%  3rd paragraph -  Our approach -> introduces the bert and gat model to solve problems like the one presented in 2nd paragraph. Key references and high-level ideas that drive the paper: what models are currently used and for which purpose?
To address these challenges, we explore two alternative deep learning architectures for automated playtesting of Candy Crush Saga levels: (a) a Bidirectional Encoder Representation from Transformers (BERT)~\cite{devlin2019bert} and (b) a Graph Attention Network (GAT)~\cite{velivckovic2017graph}. The BERT model, commonly used in natural language processing, is adapted to interpret game states as text-based inputs, allowing for a flexible representation of board configurations. This approach enables the model to handle any board size while eliminating the need for code modifications when new elements are introduced. By comparison, the GAT-based model utilizes a graph-based representation of game levels, effectively capturing relational dependencies between game elements and enabling a more flexible representation of connections between them. Unlike CNNs, the GAT architecture can learn to leverage the more complex aspects of the level topology and adapt to varying spatial structures due to the integration of relevant connections between game elements~\cite{keller2023imagesconnectionsdqngnns}.

% 4rd paragraph - What we do and how we evaluate the model using the difficulty as metric following the CNN paper and compare the three methods.
In this study, we compare the performance of CNN, BERT, and GAT-based playtesting architectures in predicting player moves and assessing level difficulty across various game levels. We train all models on historical gameplay data and assess their performance based on move prediction accuracy and level difficulty prediction, where difficulty is defined as the average Attempts Per Success (APS) of players on a given level~\cite{gudmundsson2018human}.
% Additionally, we analyze their performance in levels with and without certain connections in the graph, such as portals, that affect game topology and level complexity. 

\subsection{About Candy Crush Saga}
Candy Crush Saga (CCS) is a match-3 game structured into sequential levels, where completing one level unlocks the next level (i.e., completing level 1 unlocks level 2, etc). As of now, the game has more than 18,000 levels and features complex spatial relationships such as evolving mechanics and nonadjacent connections that challenge traditional grid-based CNN methods, serving as a valuable testbed for evaluating flexible model architectures in automated playtesting.

Each level presents the player with a two-dimensional board composed of tiles that may either be empty or contain candies and other game elements. To make a move, the player swaps two (typically adjacent) candies. However, not all swaps are valid. A move is considered valid (legal) if it either (a) results in a specific pattern of matching candies of the same color (e.g., three or more in a row, four in a square, five in a T-shape, etc.), or (b) involves a swap between two special candies.

\begin{figure}
    \centering
    \includegraphics[width=0.6\linewidth]{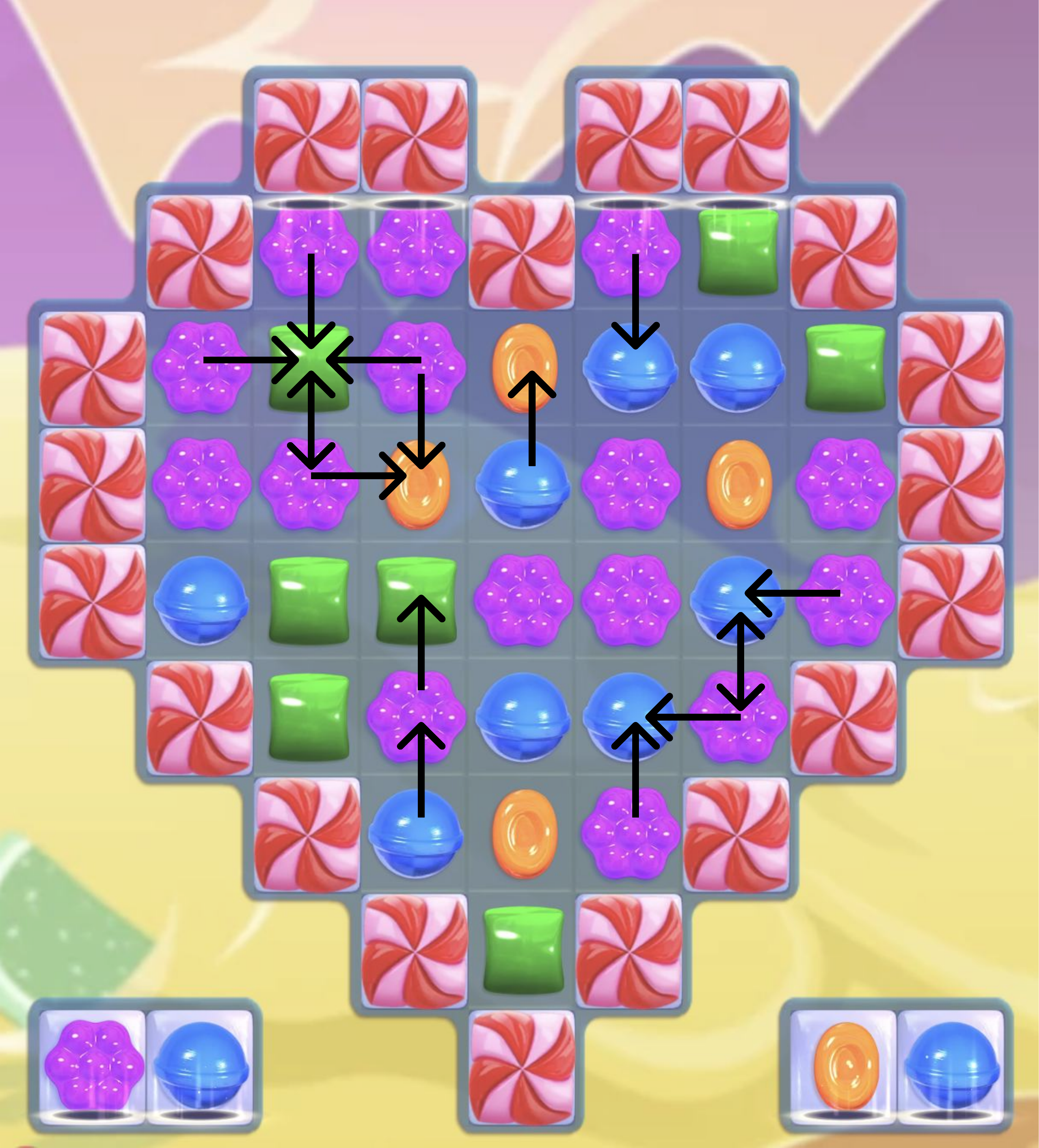}
    \caption{An example of legal moves in Candy Crush Saga is shown with black arrows on the game board.}
    \label{fig:legal_moves}
\end{figure}

When a valid move is made, the matched candies disappear. If special candies are part of the match, additional candies across the board may also be destroyed (e.g., entire rows or columns). The emptied tiles are then refilled by items cascading from above or by newly generated candies. The game client continues resolving any subsequent combos that emerge as a result of the refill. The player must wait for the board to stabilize — meaning no more candies are falling and no elements are in motion — before making the next move. Figure~\ref{fig:legal_moves} shows an example of a Candy Crush Saga game board, containing various candies and other game elements, where all legal moves are highlighted with black arrows.

\subsection{Main Contributions}
% The main contribution of this paper is ....
% Explicitly, the contributions of the paper are three-fold:
% \begin{enumerate}
%     \item a GAT architecture for training agents that can play the games at hand like human players
%     \item 
%     \item extensive experimental evaluations that validate the effectiveness of BERT, CNN and GAT on match-3 games
% \end{enumerate}

In this paper, we introduce two general-purpose automated playtesting approaches for puzzle games, addressing key limitations of CNN-based models in adapting to new game elements and modeling complex relations between distant tiles. Our main contributions can be summarized as follows:

\begin{enumerate}
    \item We examine two \textbf{BERT} models for automated playtesting: a \textit{text-based} variant that encodes board states as text sequences, allowing for a flexible and low-maintenance representation, and a \textit{board-based} variant that preserves spatial relationships while leveraging transformer architectures.
    \item We propose two \textbf{GAT} models which represent the game board in a graph structure: a \textit{GAT} model that explicitly captures relational dependencies in game boards, effectively modeling complex interactions such as portal connections, and a \textit{GAT2Edges} model that simplifies the structure by predicting move probabilities directly from edge embeddings.
    \item We conduct an extensive comparative evaluation of all proposed architectures against a CNN baseline, assessing their performance in \textit{move prediction accuracy} and \textit{level difficulty estimation} using Attempts Per Success (APS), demonstrating the advantages of structured and relational representations.
\end{enumerate}

% \begin{itemize}
% \item We identify the core limitations of CNN-based automated playtesting methods in grid-based puzzle games mainly in adapting to new game elements and modeling complex relations between distant tiles.  
% \item We examine two alternative deep learning architectures for automated playtesting of Candy Crush Saga:  
%   \begin{itemize}
%     \item A \textbf{BERT-based} model in which the board states are represented as text sequences, allowing for low maintenance costs and easy additions of the features.
%     \item A \textbf{GAT-based} model in which the board is represented in a graph structure. While it is less modular, it offers greater flexibility than CNN alternatives for encoding parts of the game board.
%   \end{itemize}
% \item We provide dataset of player state-action pairs and conduct simulation-based experiments in order to compare the performance of different architectures, measuring:
%   \begin{itemize}
%     \item \emph{Move prediction accuracy} to evaluate the ability of each model to mimic the behavior of the player.
%     \item \emph{Level difficulty prediction} using Attempts Per Success (APS), reflecting how well each model anticipates aggregated player difficulty.
%   \end{itemize}
% \end{itemize}

The rest of the paper is organized as follows. In Section~\ref{sec:related-works}, we review the related work in automated playtesting, transformer-based models in games, and graph attention networks. Section~\ref{sec:methodology} provides a comprehensive description of the configurations for each of the models. Section~\ref{sec:results} presents the evaluation metrics and a comparative analysis of the results. Concluding the paper, Section~\ref{sec:conclusion} offers a summary of the contributions and explores potential next steps.

%% file: src/1_related_works.tex
\textbf{Automated Playtesting} aims to evaluate the player's interaction with the game. Early approaches focused on building decision-making models to mimic different types of players~\cite{holmgard2016personas}. Other approaches search the action space using Monte Carlo Tree Search (MCTS)~\cite{horn2018monte} or combine MCTS with player-based heuristics to emulate specific player styles~\cite{holmgard2019evolvedmcts}. Although these methods provide structured ways to explore the game space, newer approaches leverage player data to train Neural Network models~\cite{gudmundsson2018human, clark15cnn, shao2016cnn}. Collecting extensive and diverse player information remains a major obstacle, prompting the development of a hybrid system that pairs Reinforcement Learning with real player data to simulate gameplay even on unseen levels~\cite{woillemont2022automated, shin2020_rl}.

\textbf{Transformer Models in Games} have recently emerged as powerful tools for building flexible agents that can generalize across multiple tasks. AlphaStar~\cite{vinyals2019grandmaster} is one of the early success stories of Transformers in gaming, capable of handling multiple states and thus reaching the grandmaster level in StarCraft II.  More recently, transformers have been explored for games, with the Chess Transformer~\cite{noever2020chess} demonstrating that transformer-based architectures can learn strategic play patterns. Multi-Game Decision Transformers~\cite{lee2022multi} have further extended this idea, applying transformers to play Atari games, learning policies from offline datasets in a reinforcement learning setting.
% Following works modify vanilla Transformers to enhance memory stability and replace transitional Recurrent Neural Network techniques \cite{parisotto2020stabilizing, xu2020deep}. In addition, BERT~\cite{devlin2019bert}, originally developed for natural language processing, has been adapted in our experiments to capture complex board representations, further showcasing the versatility of Transformer architectures for game playtesting.

\textbf{Graph Attention Networks (GATs)} ~\cite{velivckovic2017graph}, while not specific to gaming, have been hugely influential across domains~\cite{wu2020comprehensive}. GATs extended graph neural networks by incorporating attention on graph nodes, which allowed dynamic weighting of neighbor influences. This finds much use in playtesting, as observation and action space can often be represented in a graph structure. Traditional deep learning models, such as CNNs, struggle with long-range interactions, as observed in games like Chess and Hex, where distant board positions influence decision-making~\cite{keller2023imagesconnectionsdqngnns}, while GATs overcome this limitation by leveraging graph-based representations that allow flexible, task-specific structures, and relational inductive biases. 
% Knowledge graphs have previously been used in the context of RL. \cite{ammanabrolu2020graph} propose playing text adventure games by representing text observations as a knowledge graph and show that graph-structured text significantly increases the likelihood of being published.
% Other recent works use knowledge graphs as state representations that are learned during exploration.
Knowledge graphs have also been used in the context of RL for text adventure games, where~\cite{ammanabrolu2020graph} represent text observations as graphs and show that graph-structured representations improve agent performance.
% Many other works focus on the domain of multi-agent systems. \cite{iqbal2019actor, liu2020g2anet} learn an abstraction based on attention, which determines the transfer of knowledge between agents, while \cite{xu2022concnet} shows that previous methods fail when applied to many games and thus additionally proposes a pruning mechanism before attention. 

%% file: src/2_methodology.tex
% Methodology

In this section, we present the ML models tested in our playtesting and their respective architectures. Specifically, we explore two alternative deep learning approaches for automated playtesting, BERT and Graph Attention Networks (GAT), and compare them to the CNN-based agent proposed by~\cite{gudmundsson2018human}, which serves as our benchmark. Following~\cite{gudmundsson2018human}, all models are trained under a supervised learning framework using Cross-Entropy loss to align predictions with player actions in historical gameplay data, enabling the reproduction of human-like moves and the estimation of level difficulty. In the remaining subsections, we describe in detail the input representations, the network architectures, and the output layers for each model.

\begin{figure}[!ht]
    \centering
    \includegraphics[width=0.8\linewidth]{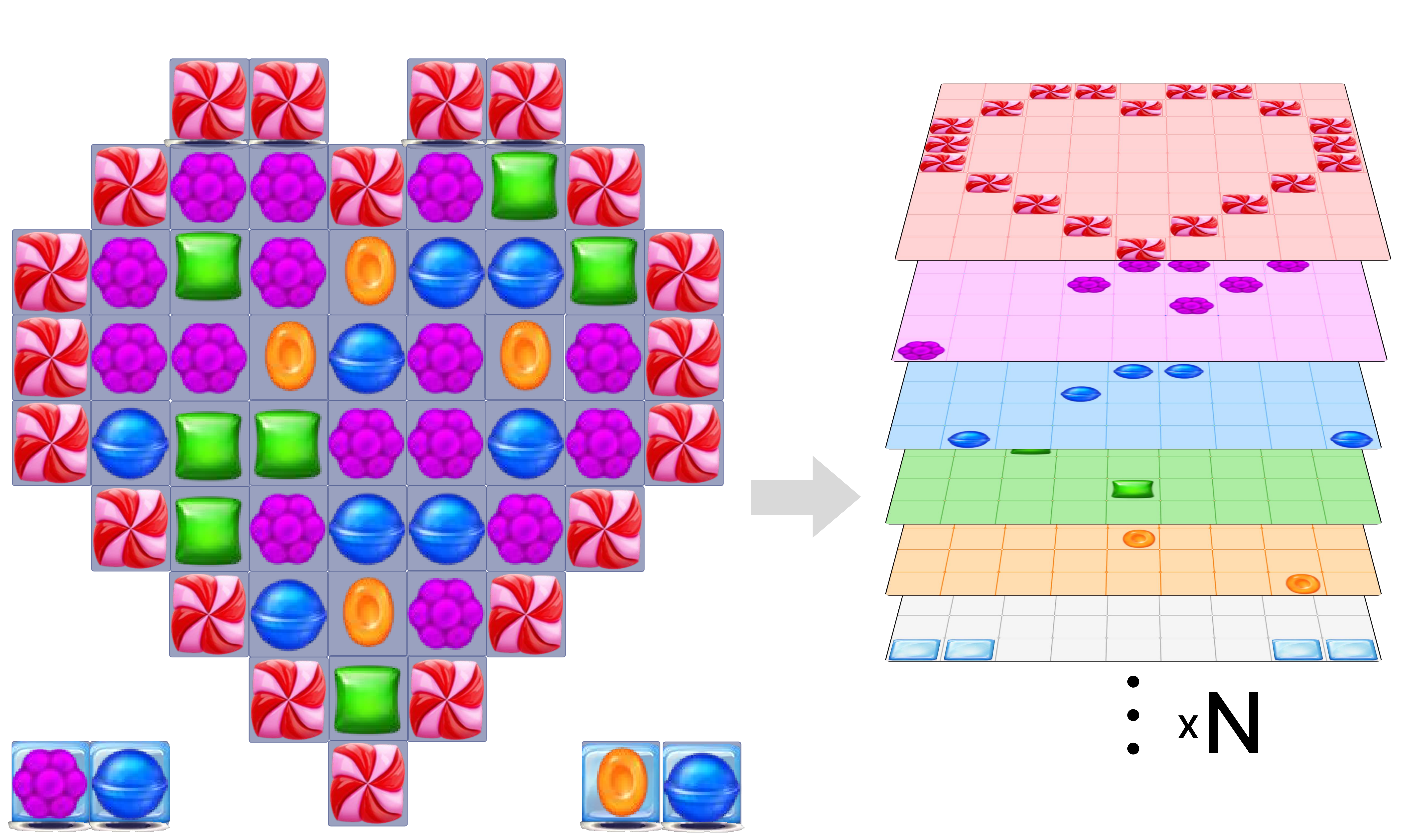}
    \caption{CNN Input. Figure adapted from \cite{gudmundsson2018human}.}
    \label{fig:cnn_input}
\end{figure}

\subsection{CNN Model}
\label{cnn_model}

The CNN model represents the game board as a 9×9×N tensor, as shown in Figure~\ref{fig:cnn_input}, where each channel encodes information derived from the board state and the level objectives. Specifically, the N channels include:

\begin{itemize} \item \textbf{Game Elements:} Each game element is encoded in a dedicated channel using a one-hot representation to indicate its
presence. \item \textbf{Level Objectives:} Each level objective is represented in a separate channel, where a float
value indicates to what extent each tile contributes to
achieving the objective (e.g., the number of collectible
jellies on the tile). \item \textbf{Legal Moves:} Legal moves show the set of valid moves a player is allowed to make given the current board state (e.g., a swap move leading to a match of three or more candies or a special combination of candies). \item \textbf{Moves Left:} Another channel includes a single number representing the remaining moves available to complete the objective of the game. \end{itemize}
% This representation includes all possible board items, game objectives, and legal moves, ensuring the model learns from valid game actions. 
Following the AlphaGo convolutional model \cite{Silver2016}, the CNN processes the board through multiple convolutional layers to extract spatial patterns. The output layer predicts a probability distribution over 144 possible actions, corresponding to all vertical and horizontal tile swaps.

\subsection{BERT Model}
We tested two BERT-based architectures as playtesting agents: a \textbf{text-based BERT model} leveraging natural-language processing and a \textbf{board-based BERT model} utilizing matrix representation. The text-based model allows for a flexible game state representation and for generalization across different game environments, while the board-based variant serves as a baseline for comparison.

\begin{figure}[!ht]

    \centering
    \begin{subfigure}{\linewidth}
        \centering
        \includegraphics[width=0.99\linewidth]{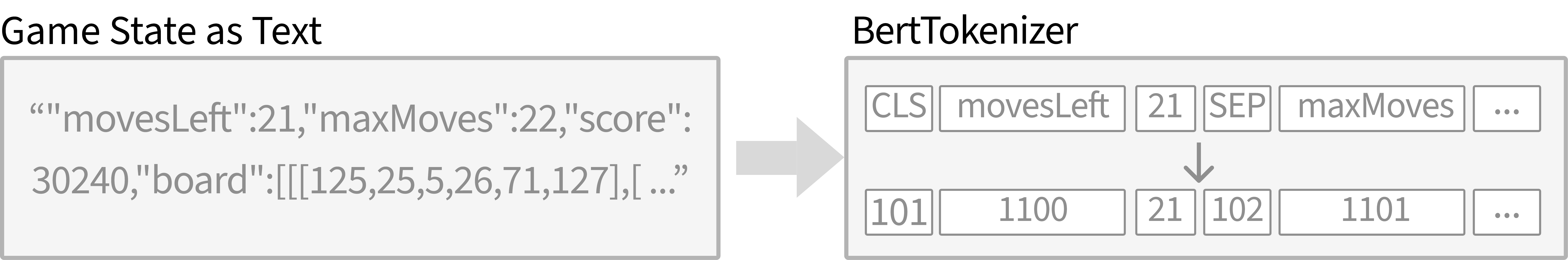}
        \caption{Text-Based Input}
        \label{fig:text_input}
    \end{subfigure}

    \par\vspace{0.5cm} % Space between figures
    
    \centering
    \begin{subfigure}{\linewidth}
        \centering
        \includegraphics[width=0.99\linewidth]{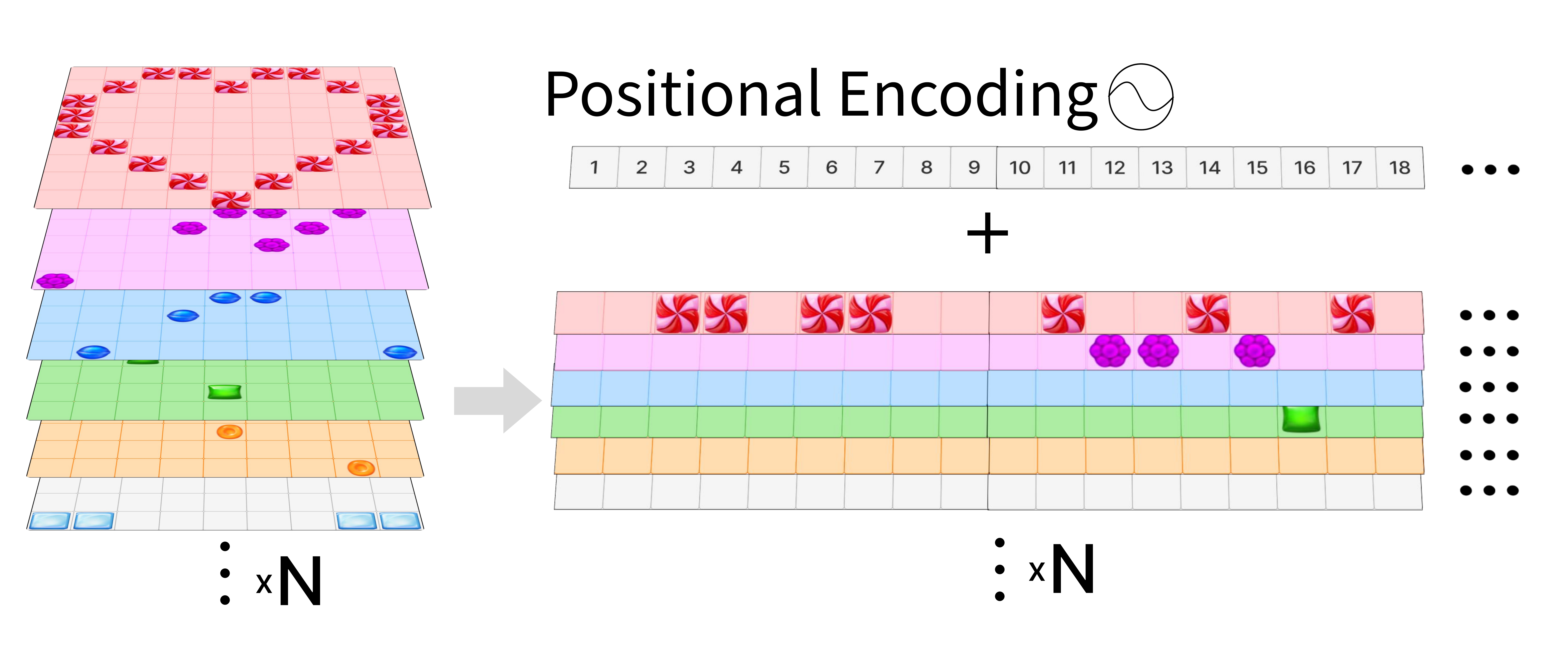}
        \caption{Board-Based Input}
        \label{fig:board_input}
    \end{subfigure}

    \caption{Input Representations of the BERT Models for the game Candy Crush Saga.}
    \label{fig:bert_input}
\end{figure}

\subsubsection{\textbf{Input Representation}}
The input to the text-based BERT model consists of the game state as text, which includes information like the board configuration, legal moves, game objectives, and number of remaining moves, as mentioned in subsection~\ref{cnn_model}. This textual representation is tokenized using BERT's built-in tokenizer, as shown in Figure \ref{fig:text_input}.

In the board-based variant, we use the same matrix representation as in the CNN method. As illustrated in Figure~\ref{fig:board_input}, the matrix is unraveled and consists of the game elements, the objectives of the level, the legal moves, and the number of remaining moves. Additionally, a trainable positional encoding is incorporated to preserve the spatial arrangement of the board elements.

\begin{figure}[!ht]
    \centering
    \includegraphics[width=0.99\linewidth]{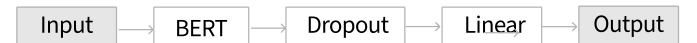}
    \caption{BERT Models' Architecture}
    \label{fig:bert_models}
\end{figure}

\begin{figure*}[!ht]
    \centering    
    \includegraphics[width=\textwidth,keepaspectratio]{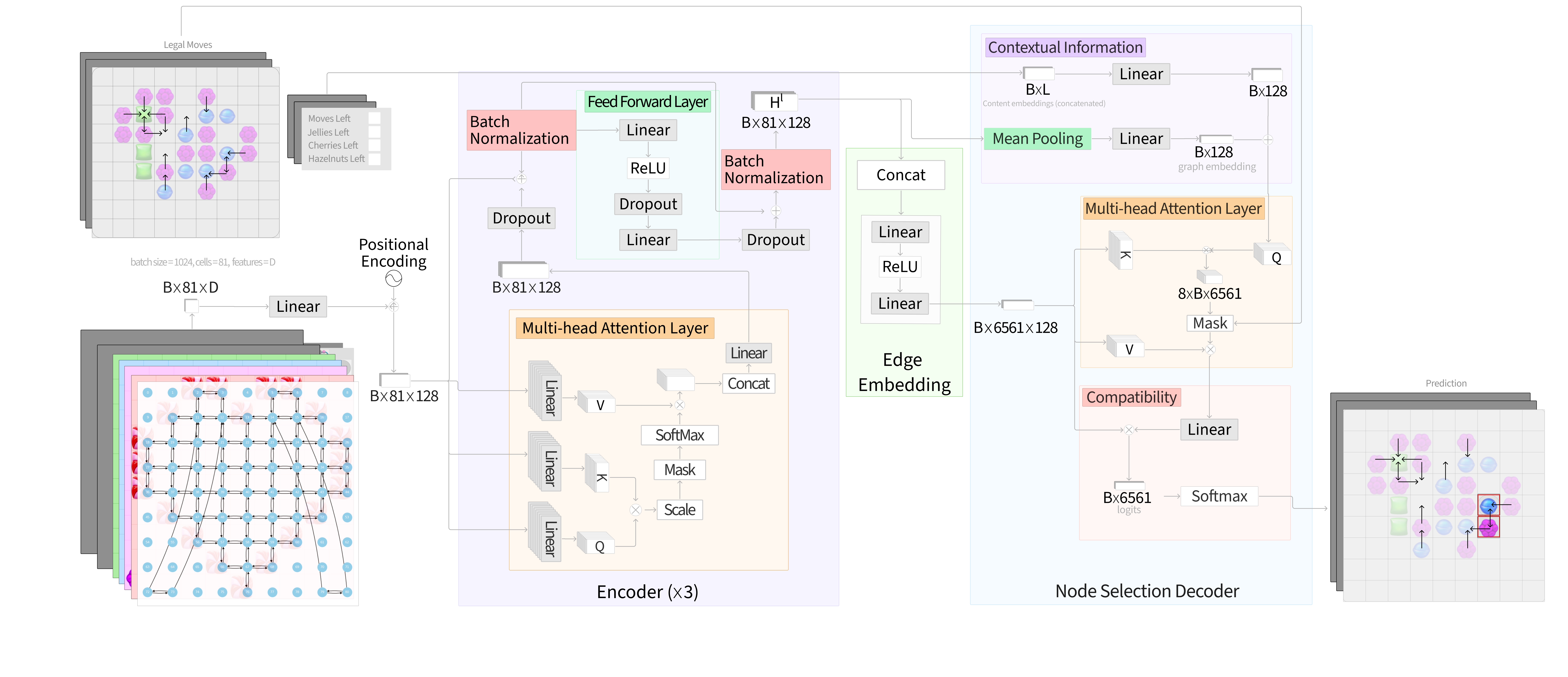}
    \caption{GAT Encoder-Decoder Architecture}
    \label{fig:gat_model}
\end{figure*}

\subsubsection{\textbf{Network Architecture}}
Figure~\ref{fig:bert_models} illustrates the architecture of both BERT model variants. The transformer-based encoder processes the input through multiple self-attention layers, enabling the model to capture dependencies between the embeddings and structural relationships within the game state.

Following the encoding stage, a dropout layer and a fully connected linear layer map the BERT output to the same action space as the CNN, producing probability distributions over 144 possible swap moves. This final output determines the likelihood of each action being selected, allowing the model to recommend optimal moves based on its learned representation of the game state.

\subsection{GAT Model}
\label{GATMODEL}
% Graph Attention Networks are a class of neural networks designed to process graph-structured data. They leverage self-attention mechanisms to dynamically assign significance to neighboring nodes, making them a suitable choice for modeling structured game states. Unlike Convolutional Neural Networks, which assume a fixed grid structure and emphasize local patterns, GATs are well-suited for capturing complex, non-local dependencies inherent in strategic board games  \cite{keller2023imagesconnectionsdqngnns}. 

We explore the representation of game states as graphs and introduce the design of our Graph Attention Network (GAT) architecture inspired by~\cite{kool2018attention}, which leverages attention mechanisms to enhance decision-making in structured environments. To investigate the trade-offs between model complexity and performance in automated playtesting, we propose two variations: the standard \textbf{GAT} model, which employs both an encoder and a decoder to integrate additional game state information, and the \textbf{GAT2Edges} variant, which omits the decoder and directly computes move probabilities from edge embeddings.

\subsubsection{\textbf{Input Representation}}
\label{input_gat}
The input to the GAT model is a graph representation of the game board, defined as $ G = (V, E) $, where $ V $ represents the nodes and $ E $ the edges.
\begin{itemize}
\item \textbf{Nodes $ V $:} the set of all nodes in the graph, each corresponding to a tile on the board and characterized by a number of features that describe it.
\item \textbf{Edges $E$:} relationships between pairs of tiles which define the graph structure, such as adjacency in the board (left, right, up, down) or direct connections between tiles through, e.g. portals.
\item \textbf{Node Features:} Each tile is associated with a feature vector that includes a one-hot encoding of all game elements, as well as the level objectives. For the \textbf{GAT2EDGES} model - for which we disregard the decoder part - we also include the goal entities and moves left as described in subsection~\ref{cnn_model}, so that both GAT variants take into account the same information.
\item \textbf{Positional Encoding:} We incorporate a trainable positional encoding into each node's feature vector to capture spatial dependencies.
\end{itemize}

\subsubsection{\textbf{Encoder}}
The encoder processes the graph $ G = (V, E) $, as defined earlier, to capture the level topology and output informed representations of the node embeddings. Each encoder layer consists of two sublayers, a multi-head attention sublayer and a feed-forward sublayer, while dropout and batch normalization are applied after each of the aforementioned sublayers to enhance generalization and stability~\cite{velivckovic2017graph, vaswani2017attention}.  

The multi-head attention layer includes self-attention to dynamically weigh the importance between the connected nodes, e.g. nodes with the same candies have higher attention weights. By stacking three encoder layers ($\times$3), we aim to propagate information learned in previous layers, resulting in richer node embeddings. These embeddings incorporate not only information from immediate neighbors but also from their extended neighborhoods, reaching up to three connections away in the graph.

\subsubsection{\textbf{Edge Embeddings}}
The encoder outputs enriched node embeddings, capturing both local and long-range dependencies within the game board. To model all possible moves, we construct edge embeddings by concatenating the embeddings of each node pair. Depending on the node pair, the move may indicate a double-tap if the concatenated nodes are the same, a swap if the concatenated nodes are neighboring in the grid, or a special action (e.g. frog) if the concatenated nodes are further away on the board. This approach expands the action space beyond the 144 typical moves of swap-based models, evaluating every possible unidirectional tile combination, resulting in a total of $
\frac{(9 \times 9)\left((9 \times 9)+1\right)}{2} = 3321$ possible moves. The constructed edge embeddings are then processed through a feedforward layer with a ReLU activation function, allowing the model to capture relational patterns between tiles more effectively. This transformation plays a crucial role in both model variants: in the standard GAT model, these enriched edge embeddings contribute to the decoder’s decision-making process, while in GAT2Edges, where the decoder is omitted, they serve as the direct input for computing move probabilities.

\subsubsection{\textbf{Decoder}}
The decoder receives the output of the encoder and the edge embeddings, along with additional inputs, namely the legal moves, the moves left, and the game objectives as described in Subsection~\ref{cnn_model}. The decoder is structured into three main components: the Contextual Information, the Multi-head Attention, and the Compatibility sublayers~\cite{kool2018attention}. 

\begin{itemize}
\item  \textbf{Contextual Information Sublayer}: This sublayer constructs the decoder's query (denoted as Q in Figure~\ref{fig:gat_model}) by combining the game objectives, the remaining moves, and the graph embedding, the latter of which is obtained through mean pooling over the node embeddings. The output is a representation that encapsulates the learned representation of the current board state and the contribution of different nodes toward achieving the game objective. This effectively generates a summary of the state for which a prediction is made.

\item \textbf{Multi-head Attention sublayer}: This sublayer uses the cross-attention mechanism as the query differs from the key and the value. The query is generated from the contextual information sublayer, while the key and value correspond to the edge embeddings, which indicate all possible moves in the graph. We then get the compatibility between an action (edge embedding) and the goal of the game (query), while we mask the non-legal moves so to limit the model to only legal moves.

\item \textbf{Compatibility Sublayer}: This sublayer compares the actions (edge embeddings) with the output of the multi-head attention layer, which indicates how much an action contributes to the objective of the game, and gives a probability out of all possible moves on the board.

\end{itemize}

%% file: src/3_results.tex
% Experiments
% - We compare CNN results to the GAT and BERT results for CCS and additionally show prediction accuracy.

In this section, we compare the performance of the CNN, BERT, and GAT models in predicting player moves and estimating level difficulty in CCS. We assess move prediction based on accuracy to capture how often the model’s predicted action aligns with human gameplay data. For difficulty estimation, we perform playtesting on the entire level and use Attempts Per Success (APS) as a metric to compare model predictions with actual player performance. The following subsections provide a detailed overview of the dataset used, the experimental setup, and the comparative analysis of the models.

\subsection{Dataset}
% - How many in each set and distribution among game modes and difficult levels. Balanced based on game mode and ordinal levels.

We construct our dataset from historical gameplay data collected from CCS to estimate move prediction accuracy and level difficulty. For each of these evaluation tasks, we use separate data subsets to ensure a proper and unbiased evaluation. The first task focuses on predicting player moves based on a board state, while the second evaluates model performance by simulating complete level playthroughs under realistic playtesting conditions.

For move prediction, we use player state–move data and split it into training, validation, and test sets to evaluate model performance on unseen levels and assess its generalization capability. The dataset is balanced across different game modes and progression levels. Specifically, we collect approximately $\mathrel{\sim}\!400$K samples per game mode across the 10 game modes available in the game. The training data consists of approximately $33\%$, $36\%$, and $31\%$, easy, medium, and hard levels respectively. To further prevent data leakage, the test set is drawn from a non-overlapping time window and consists of $\sim1$M samples across all game modes.

To evaluate level difficulty, we conduct inference-based simulations on a separate set of $\sim300$ levels, uniformly sampled across levels 1000 to 15,000, covering 10 different game modes. Each level is playtested using 1,000 game rounds per model, allowing us to evaluate the difficulty of each level, and compare model and player performance.

\subsection{Experimental Setup}

All proposed models were trained using the Adam optimizer, weight regularization of $1e-9$, dropout probability of $0.1$, and batch size of $128$. A learning rate of $1e-4$ was used for all GAT models and the board-based BERT, while text-based BERT and CNN were trained with a learning rate of $1e-3$. The learning rate decayed at a rate of $0.98$ every $25k$ steps. Training is conducted for 10 epochs on an \textit{NVIDIA P100} GPU and lasted about $5$, $26$, $40$, $34$ and $17$ hours for the CNN, Board-based BERT, Text-based BERT, GAT and GAT2EDGES model, respectively.

Inference-based simulations are executed using Kubernetes with 6 replicas, each allocated 8 CPUs, totaling 48 CPUs running in parallel.
Simulation time reveals that GAT-based models exhibit higher computational demands compared to CNNs, with GAT averaging 5 minutes per inference—twice the duration of CNN (2.5 minutes)—while GAT2Edges achieves the lowest inference time at 2.4 minutes. 

% In terms of resource utilization, CNN remains the most computationally efficient model, with an average CPU usage of 3.37\% and memory consumption of ~1.93GB, whereas GAT requires ~15.34\% CPU and $\sim6.12$GB of memory, reflecting the increased complexity of graph-based processing. GAT2Edges provides a trade-off, reducing CPU usage to 7.60\% and memory consumption to 5.58GB, making it a more resource-efficient alternative while still leveraging a graph-based representation.

The models vary significantly in the number of trainable parameters, as shown in Table~\ref{tab:test_accuracy}. The BERT-based models have the highest number of parameters due to their transformer-based architecture, making them the most computationally demanding. The CNN model has a moderate number of parameters, primarily driven by its convolutional and residual blocks. Among the graph-based models, GAT has a higher parameter count than GAT2Edges, as the latter omits the decoder component, reducing complexity while still leveraging graph-based processing.

To quantify computational costs, we analyze floating point operations (FLOPs). The BERT board-based model exhibits the highest FLOPs, while graph-based models require significantly fewer FLOPs, with GAT2Edges being the most efficient among them. The variation between text- and board-based FLOPs in BERT is due to different token lengths. Compressing tokens before feeding them to the attention layers could improve efficiency and should be explored in a future iteration. The CNN model is the most computationally efficient overall, benefiting from optimized convolutional operations.

\subsection{Model Comparison}

We evaluate the performance of the models along the following dimensions: \textbf{(1)} move prediction accuracy and \textbf{(2)} level difficulty estimation. 
Additionally, we train an alternative variant for each graph-based model, GAT and GAT2EDGES, where portal information is omitted meaning the model does not receive portal connections as edges in the input graph. The goal is to validate the importance of modeling the input as a graph and the significance of incorporating topology-aware information into that input. Portal connections cannot be represented using the CNN input as those are connections between non-adjacent tiles, further motivating the need for graph-based approaches.

\begin{table}[h]
    \centering
    % \captionsetup{justification=centering, labelsep=newline, font=small} % Center caption
    % \caption{Move Prediction Accuracy on test set (Top-1, Top-2, Top-3, Top-6, and Top-12) for the three models presented and their respective variants. The results clearly show that the GAT model outperforms all alternative approaches across all accuracy metrics.}
    \caption{The accuracy of move predictions on the test set (Top-1, Top-2, Top-3 and Top-6) for CNN, BERT, and GAT-based models together with the number of parameters and floating point operations (FLOPs) in millions. "NP" denotes the variants trained without portal connections in the graph. Results highlight that GAT outperforms all alternative approaches.}
    \renewcommand{\arraystretch}{1.2}
    \setlength{\tabcolsep}{3pt} % Adjust column spacing
    \resizebox{0.49\textwidth}{!}{ 
    \begin{tabular}{
    l
    p{1cm}
    p{1cm}
    p{1cm}
    p{1cm}
    |r
    r
    }
        \toprule
        \textbf{Model} & \textbf{Top-1} & \textbf{Top-2} & \textbf{Top-3} & \textbf{Top-6} & \textbf{\scalebox{0.9}{Params}} & \textbf{\scalebox{0.9}{FLOPs}}\\
        \midrule
        
        % \multicolumn{7}{c}{\textbf{Test accuarcy on all levels}} \\
        % \hline\hline 
        
        CNN & 0.5333 & 0.7241 & 0.8233 & 0.9431 & 2.75 & 433 \\
        \midrule
        BERT Text-Based  & 0.2220 & 0.3639 & 0.4583 & 0.6057 & 4.40 & 686\\
        BERT Board-Based & 0.3584 & 0.5387 & 0.6586 & 0.8497 & 10.34 & 1723\\
        
        \midrule
         GAT & \textbf{0.5437} & \textbf{0.7341} & \textbf{0.8315} & \textbf{0.9477} & \multirow{2}{*}{0.84} &  \multirow{2}{*}{1365}\\
        GAT-NP & 0.5430 & 0.7332 & 0.8303 & 0.9471 \\
        
        % \midrule
        GAT2EDGES & 0.5240 & 0.7166 & 0.8182 & 0.9422 & \multirow{2}{*}{0.70} & \multirow{2}{*}{934} \\
        GAT2EDGES-NP & 0.5246 & 0.7174 & 0.8188 & 0.9425 \\

        % \midrule
        % \multicolumn{7}{c}{\textbf{Models trained on difficult levels only}} \\
        % \hline\hline
        % CNN &  & 0.4943 & 0.6869 & 0.7914 & 0.9253 & 0.9812 \\
        % \midrule
        % \multirow{2}{*}{GAT2EDGES} & Encoder only & 0.5094 & 0.7014 & 0.8043 & 0.9344 & 0.9904 \\
        % & w/o portal connections & 0.5103 & 0.7025 & 0.8053 & 0.9350 & 0.9906 \\
        % \midrule
        % \multirow{2}{*}{GAT} & Encoder-Decoder & \textbf{0.5249} & \textbf{0.7153} & \textbf{0.8150} & \textbf{0.9390} & 0.9913 \\
        % & w/o portal connections & 0.5222 & 0.7135 & 0.8137 & 0.9387 & \textbf{0.9914} \\
        \bottomrule
    \end{tabular}
}
    \label{tab:test_accuracy}
\end{table}
\begin{figure*}[!ht]
    \centering
    \includegraphics[width=0.99\linewidth]{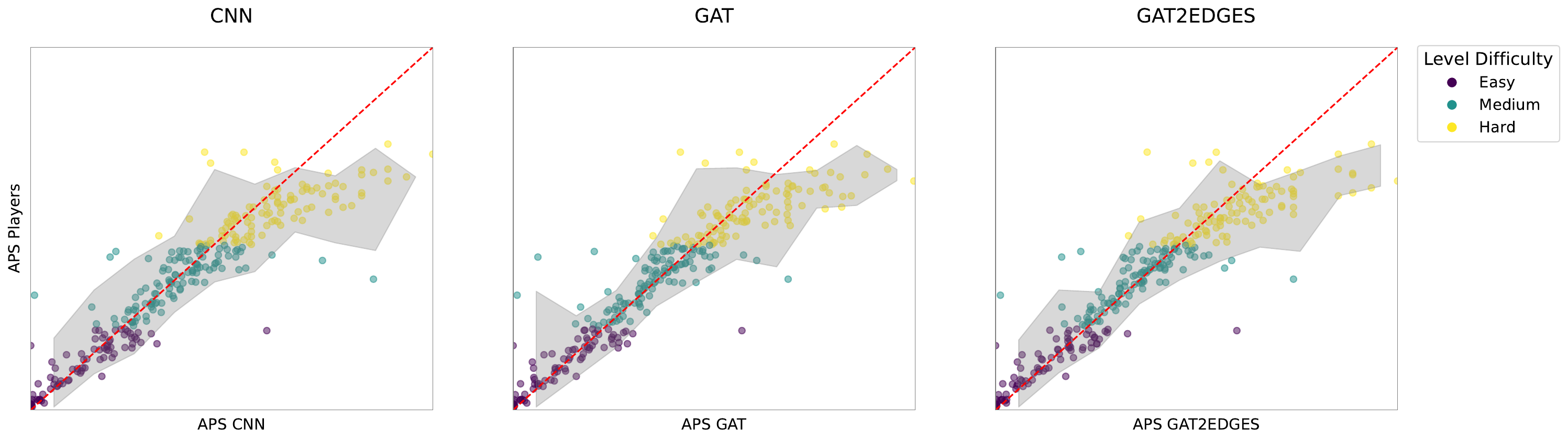}
    \caption{Correlation between observed Attempts Per Success (APS) from player data and APS values predicted by each model. All models exhibit similar predictive trends, with graph-based models demonstrating slightly lower variance.}
    \label{fig:correlation_plot}
\end{figure*}

\subsubsection{\textbf{Move Prediction Accuracy}}
The performance of each model in predicting player moves is reported in Table~\ref{tab:test_accuracy}, where top-k accuracy is measured on the test dataset. Specifically, top-1, top-2, and top-3 accuracy indicate whether the actual player move appears within the first k predictions made by the model. The higher the accuracy values are, the better the model is aligned with player decision-making patterns.

As shown in Table~\ref{tab:test_accuracy}, the proposed GAT approach shows better performance compared to convolution- and transformer-based alternatives in predicting player moves. Out of the models presented, text-based Bert's performance was significantly lower, hinting that structuring the input data can provide accuracy gains. The CNN and GAT-based models perform quite similarly; with the CNN performing somewhere in-between GAT and GAT2EDGES in all metrics.

The full GAT model benefits from the added complexity and features, outperforming the GAT2EDGES approach, which simply performs edge prediction given the input graph representation. Interestingly, while the GAT2EDGES model achieves higher validation accuracy when portal connections are omitted, the full GAT model exhibits a slight accuracy improvement when these connections are included. These findings suggest that with the additional parameters and metadata, portal information can indeed enhance the prediction of the models.

\subsubsection{\textbf{Level Difficulty Estimation}}

While accuracy provides a good guideline, our primary interest lies in how the models perform during simulations, since those indicate whether the models can capture player behavior and accurately measure level difficulty. As the primary difficulty metric, we use Attempts Per Success (APS), where APS is defined as the ratio of total game rounds played to the number of successful completions. A higher APS value indicates a more challenging level.

To perform the simulations, we employ the trained models (i.e. agents) as policies to evaluate all possible actions \( a \in A \) given a game state \( s \), where \( A \) represents the set of available actions. At each step, the agent selects and executes the action with the highest probability: $\max_{a \in A} P(a | s)$. The environment then transitions to a new state \( s' \) with an updated set of available actions \( A' \), and the agent re-evaluates the possible actions based on the new state. This iterative process continues until a terminal state is reached, either by fulfilling the game objective (win) or exhausting all available moves (loss). This process is repeated for each game round, and by iterating multiple times, we obtain a meaningful measure of the level's overall difficulty by calculating the APS value.

We perform these simulations on the inference dataset and summarize the results in Table~\ref{tab:mae_results_no_mean}. Specifically, we report the median absolute error (MedianAE) between the predicted and actual APS values, as it provides a robust measure that is less sensitive to outliers. Additionally, we include the Standard Deviation (STDDEV) of the absolute error to get insights into the variability of predictions. To account for differences in difficulty scaling, we present results separately for different level segments, as higher APS values correspond to more challenging levels, and thus leading to proportionally larger absolute errors. We also include the aggregated metrics across all segments to give an overall picture of the different models. Finally, we further compute the metrics only for the levels with portals to better understand the value of portal connections.

\begin{table}[h]
    \centering
    % \caption{Median Absolute Error (MedianAE) and Standard Deviation (STDDEV) of Attempts per Success (APS) for each model grouped by difficulty.}
    \caption{Median Absolute Error (MedianAE) and Standard Deviation (STDDEV) of the absolute error between model and player APS, grouped by level difficulty. Graph-based models consistently show superior performance, with simpler architectures performing better on easier levels and the full GAT model excelling in more challenging scenarios.}
    \renewcommand{\arraystretch}{1.2}
    \setlength{\tabcolsep}{5pt} % Adjust column spacing
    \resizebox{0.49\textwidth}{!}{
    \begin{tabular}{llcc cc}
        \toprule
        \multirow{2}{*}{\textbf{Difficulty}} & \multirow{2}{*}{\textbf{Model}} &
        \multicolumn{2}{c}{\textbf{All Levels}} &
        \multicolumn{2}{c}{\textbf{Portal Levels}} \\
        \cmidrule(lr){3-4} \cmidrule(lr){5-6}
         & & \textbf{MedianAE} & \textbf{STDDEV} & \textbf{MedianAE} & \textbf{STDDEV} \\
        \midrule
        
        \multirow{4}{*}{Easy}  
        & CNN             & 0.23 & 3.80  & 0.25  & \underline{0.55} \\
        & GAT             & \underline{0.23}  & \underline{3.35}  & \underline{0.17}  & \underline{0.55} \\
        & GAT-NP          & 0.25  & \textbf{2.73}  & 0.18  & 0.69 \\
        & GAT2EDGES       & \textbf{0.22}  & 4.14  & 0.20  & \underline{0.55} \\
        & GAT2EDGES-NP    & \underline{0.22}  & 3.49  & \textbf{0.16}  & \textbf{0.51} \\
        \midrule

        \multirow{4}{*}{Medium}  
        & CNN             & 1.80  & 19.66 & 2.14  & 3.70 \\
        & GAT             & \underline{1.34}  & \textbf{6.89}  & 1.85  & \underline{3.16} \\
        & GAT-NP          & 1.50  & 21.87 & \textbf{1.30}  & 3.55 \\
        & GAT2EDGES       & \textbf{1.22}  & \underline{9.33}  & \underline{1.40}  & 3.35 \\
        & GAT2EDGES-NP    & 1.40  & 10.67 & 1.54  & \textbf{2.99} \\
        \midrule

        \multirow{4}{*}{Hard}  
        & CNN             & 14.00 & 61.98 & 15.24 & \underline{83.52} \\
        & GAT             & \textbf{10.03} & \textbf{59.71} & \textbf{10.69} & \textbf{83.10} \\
        & GAT-NP          & 12.93 & 67.62 & 22.25 & 89.58 \\
        & GAT2EDGES       & \underline{12.22} & \underline{67.36} & \underline{11.17} & 88.77 \\
        & GAT2EDGES-NP    & 12.89 & 136.17 & 16.67 & 174.03 \\
        \midrule

        \multirow{1}{*}{All}  
        & CNN             & 1.98  & \underline{41.23} & 1.78  & \underline{51.36} \\
        & GAT             & 1.61  & \textbf{38.02} & 1.86  & \textbf{50.88} \\
        & GAT-NP          & 1.82  & 44.28 & \underline{1.58} & 55.14 \\
        & GAT2EDGES       & \textbf{1.55}  & 42.94 & \textbf{1.54}  & 53.72 \\
        & GAT2EDGES-NP    & \underline{1.58}  & 82.50 & 1.63  & 102.96 \\
        \bottomrule
    \end{tabular}
    }
    \label{tab:mae_results_no_mean}
\end{table}

Surprisingly, move prediction accuracy results do not directly translate to the simulation performance. In general, the graph-based models outperform the CNN baseline, yet the benefits of each graph-based model variant are nuanced for easier levels. For easy and medium levels, the encoder only (GAT2EDGES) model achieves lower median errors, while the full GAT tends to be more robust as indicated by the lower standard deviation values. In contrast, for harder levels, where capturing complex player movements and strategies is critical, the full GAT model clearly excels in achieving both lower median absolute errors and reduced variability. Additionally, incorporating portal connections in the graph input proves beneficial in the more difficult scenarios, further enhancing performance. A similar trend is observed when evaluating levels that include portals. Once again, the encoder-only GAT2EDGES model performs best for easier levels, while the full GAT model with the portal connections included consistently outperforms other architectures in more complex levels.

In addition to absolute errors, designers are interested in differentiating between levels, particularly identifying which levels are more difficult than others. One way to show this is using a scatter plot, as shown on Figure~\ref{fig:correlation_plot}, which illustrates the distance between the APS values of the players and models. We observe that the difficulty estimation of the model strongly correlates with that of the players. However, the correlation is weaker on harder levels, where players tend to win more frequently than the models predict. Additionally, this plot highlights that the CNN, GAT, and GAT2EDGES models exhibit comparable overall performance, but the graph-based models have lower variability, suggesting greater consistency in their predictions.

%% file: src/5_conclusion.tex
This study highlights the significance of adaptable and scalable automated playtesting in grid-based puzzle games, where frequent feature updates necessitate structural adjustments in how game boards are represented within models. Traditional CNN-based approaches, while effective, often struggle with these evolving dynamics due to their reliance on fixed spatial structures. In contrast, BERT and GAT architectures offer more modular and flexible representations. BERT encodes game states as token sequences, capturing sequential dependencies, while GAT explicitly models non-adjacent tile relationships, such as portal connections, in its graph structure. This ability to incorporate complex relationships makes GAT particularly well-suited for handling dynamic gameplay elements and evolving game mechanics.

Our results indicate that GAT-based models can match or surpass the performance of CNN models, while avoiding the constraints of an image-like input representation. Although BERT-based models offer greater flexibility, they do not perform as well as the structured approaches such as CNN and GAT in this domain. These findings highlight the potential of graph representations for effectively modeling puzzle game states, such as those in CCS. These results establish a strong foundation for developing more generalizable AI models that can adapt to increasingly complex level designs, paving the way for more advanced automated playtesting and AI-assisted game design.
% Taken together, these observations pave the way for multiple directions of future research to better represent the data and build new modular architectures. Future directions could further increase modularity of the GAT based alternatives or effectively improve vanila Transformer models to be able to compete with the deep learning models proposed (i.e CNN, GAT). Ultimately, the findings show that graph representations can effectively encapsulate the state of puzzle games (Candy Crush Saga) and offer a robust foundation for complex level designs. 

Future research could focus on enhancing the efficiency of GAT-based models by exploring techniques such as model pruning, optimized graph construction, and advanced computational strategies to reduce training and simulation times. Additionally, further refinement of Transformer architectures could be explored by enhancing the text-based input representation. This may improve their predictive capabilities, potentially providing a better balance between flexibility and accuracy. Investigating hybrid approaches that combine the structured advantages of GAT with the sequential processing strengths of Transformers could also be a promising direction, enabling models to adapt more effectively to the evolving complexity of puzzle game mechanics.